# Systematic investigation into generalization of COVID-19 CT deep learning models with Gabor ensemble for lung involvement scoring

# PREPRINT

**Michael James Horry[1], Subrata Chakraborty[1*], Biswajeet Pradhan[1], Maryam Fallahpoor[1], Chegeni Hossein[2], Manoranjan Paul[3]**

[1]Center for Advanced modelling and Geospatial Information Systems
School of Information, Systems & Modelling, Faculty of Engineering and IT, University of Technology Sydney.
Sydney, NSW 2007, Australia; michael.j.horry@student.uts.edu.au;
subrata.chakraborty@uts.edu.au; biswajeet.pradhan@uts.edu.au

[2]Fellowship of Interventional Radiology. Imaging Center, Mehr General Hospital, Tehran, Iran. cheginhosein@gmail.com

[3]Machine Vision and Digital Health (MaViDH), School of Computing and Mathematics, Charles Sturt University, Bathurst, NSW 2795, Australia, mpaul@csu.edu.au

## Abstract

The COVID-19 pandemic has inspired unprecedented data collection and computer vision modelling efforts worldwide, focusing on diagnosis and stratification of COVID-19 from medical images. Despite this large-scale research effort, these models have found limited practical application due in part to unproven generalization of these models beyond their source study. This study investigates the generalizability of key published models using the publicly available COVID-19 Computed Tomography data through cross dataset validation. We then assess the predictive ability of these models for COVID-19 severity using an independent new dataset that is stratified for COVID-19 lung involvement. Each inter-dataset study is performed using histogram equalization, and contrast limited adaptive histogram equalization with and without a learning Gabor filter. The study shows high variability in the generalization of models trained on these datasets due to varied sample image provenances and acquisition processes amongst other factors. We show that under certain conditions, an internally consistent dataset can generalize well to an external dataset despite structural differences between these datasets with f1 scores up to 86%. Our best performing model shows high predictive accuracy for lung involvement score for an independent dataset for which expertly labelled lung involvement stratification is available. Creating an ensemble of our best model for disease positive prediction with our best model for disease negative prediction using a min-max function resulted in a superior model for lung involvement prediction with average predictive accuracy of 75% for zero lung involvement and 96% for 75-100% lung involvement with almost linear relationship between these stratifications.

---

[*] Corresponding Author

## 1. Introduction

The COVID-19 pandemic remains a serious worldwide problem with around 2.6 million deaths attributable to this disease as of March 11, 2021 (Statista, 2021). Whilst vaccination efforts are now underway, the virus is still circulating and many countries remain locked down with new variants of the disease recently emerging (Centers for Disease Control and Prevention, 2021). Pandemics such as COVID-19 are considered by expert panels to be a recurring threat with the expectation that respiratory viruses causing severe pneumonia will continue to mutate and emerge (The Global Preparedness Monitoring Board, 2021).

From the start of this pandemic around December 2019 the research community has worked tirelessly to detect, measure, and treat COVID-19 and its devastating health and economic consequences more effectively. Data scientists and medical AI specialists have recognised the promise of automated computer vision diagnosis and stratification of COVID-19 from medical images including Computed Tomography (CT) (Butt et al., 2020; Chen et al., 2020; Gozes et al., 2020; Gunraj et al., 2020; Kumar et al., 2020; Li et al., 2020; Maghdid et al., 2020; Ni et al., 2020; Pham, 2020; Polsinelli et al., 2020; Shan et al., 2020; Silva et al., 2020; Wang et al., 2020; Zheng et al., 2020), chest X-ray (Blain et al., 2020; Civit-Masot et al., 2020; J. P. Cohen, L. Dao, et al., 2020;

El-Din Hemdan & Shouman, 2020; M. J. Horry et al., 2020; Liu et al., 2020; Oh et al., 2020; Ucar & Korkmaz, 2020; Yoo et al., 2020; J. Zhang et al., 2020) and point of care ultrasound (Born et al., 2020; Roy et al., 2020) imaging modes. These computer vision methods almost universally employ a trainable convolutional neural network (CNN) (Bengio & Lecun, 1997) as an image feature classifier, in a process commonly referred to as "Deep Learning". The CT imaging mode has received particular focus following results of a large study in the early days of the COVID-19 pandemic showing that chest CT outperformed Polymerase Chain Reaction (PCR) based lab testing in the diagnosis of COVID-19 (Ai et al.). CT imaging also has the advantage of indicating to clinicians the extent of a patient's lung involvement associated with COVID-19 pneumonia, thereby providing information about infection progression and severity (Bernheim et al., 2020) which can be used to appropriately manage the patient, and clinic resources in general, for best outcomes.

Despite this research effort, applications of computer vision assisted diagnosis and stratification of COVID-19 has not yet reached operational maturity due to concerns such as dataset bias and limited validation against external datasets (Scott & Coiera, 2020; Willemink et al., 2020) resulting in apprehension relating to the likely performance of these models in clinical settings. These concerns are suitably justified when one considers the many possible sources of systematic bias that could be introduced by differences in medical image acquisition apparatus and operating methods, or differences in patient morphology between sample and target populations (Roberts et al., 2021). Other potential sources of sample bias include differences in the down-sampling method employed to convert high resolution medical images in the form of multiple slices contained in Digital Imaging and Communications in Medicine (DICOM) files into smaller numbers (often a single slice) of lower resolution compressed images suitable for training deep learning models, due to systematically different slice selection and image compression artefacts (Roberts et al., 2021). The possibility of sample bias is further compounded when one considers that disease positive and disease negative sample sets are frequently independently sourced and processed in the publicly available CT scan data corpuses.

The reported association of COVID-19 mortality with health-care resource availability (Ji et al., 2020) highlights the fact that clinical resources are finite and limited. Clinical tools based on deep learning models for COVID-19 diagnosis, stratification and prognosis would help ensure more efficient clinical resource allocation resulting in better outcomes for patients. These benefits can only be realised if trained models perform well in real clinical situations, therefore a clear understanding of the factors affecting deep learning model generalization performance is critical to translating the immense body of research in this field into well-engineered clinical tooling.

This paper presents the results and interpretation of a systematic inter-dataset study across four publicly available COVID-19 CT image datasets and one high-quality privately curated image dataset which we have included as an unbiased data corpus of known provenance. We first closely replicate the models described in the dataset source studies using a consistently trained off-the-shelf Densenet-121 Convolution Neural Network (CNN). This creates a comparable base set of trained computer vision COVID-19 diagnostic models. Each of these models is then assessed against each other's test data partition with Histogram Equalization (OpenCV, n.d.) and Contrast Limited Adaptive Histogram Equalization (CLAHE) (Zuiderveld, 1994) image pre-processing techniques applied, these being the most commonly employed methods of reducing systematic image variance in brightness and contrast (Al-Ameen et al., 2015). Each test is repeated with the addition of a learning Gabor filter (Feichtinger & Strohmer, 2012) which has been shown by some studies to improve CNN accuracy in thoracic imaging applications (Han et al., 2014; Paulraj & Chellliah, 2018). From these tests we gain insights into the ability of each model to generalize to external datasets, along with an indication of the effects of image histogram pre-processing and application of a learning Gabor filter.

We proceed to test the predictive capability of each trained model against an independent dataset for which COVID-19 lung disease involvement has been measured and labelled by expert radiologists (Morozov et al., 2020). This final step reveals which training corpus and pre-processing options result in trained deep learning models that are best able to predict COVID-19 lung involvement scores and therefore be considered successfully generalized. Finally, we ensemble our two best models and use a min-max algorithm to achieve very high predictive scores for lung involvement against the expertly labelled control data set. Our discussion considers the source dataset attributes and provenance resulting in successful generalization and we propose a minimal set of guiding principles for the development of generalizable deep learning computer vision models for COVID-19 and other thoracic diseases.

This study represents the first systematic investigation into the generalizability of models trained on COVID-19 CT datasets. The outcome of this investigation leads us to incredibly promising and, to our knowledge, state-of-the-art results in applying a deep learning model trained on one dataset to successfully inference the lung involvement strata for a completely independent dataset. Our method for ensembling models using differentiated pre-processing techniques for disease positive and disease negative prediction is novel and foreseeably applicable to a range of computer vision applications. We analyse our results, both poor and excellent, and provide a series of novel insights into factors affecting the generalization of deep learning computer vision models. We hope that the research community can build upon our findings to create more mature deep learning models that will generalize to clinical practice.

## 2. Related Work

Although there are many published studies in the use of computer vision in the diagnosis and stratification of COVID-19 very few of these (Silva et al., 2020) investigate the inter-dataset generalization behaviour of the described models. In our review we came across only a single study that externally validated the performance of their model against an independent dataset (Silva et al., 2020). In this paper a cross dataset generalization study showed a drop in accuracy for the presented model from 87.68% to 56.16% concluding that the generalization power of the deep learning models under consideration is "far from acceptable" for a realistic scenario.

The question of deep learning generalization in the context of automated pneumonia detection from the chest X-ray imaging mode was comprehensively investigated in (Zech et al., 2018). This study trained and evaluated a CNN-based deep learning model for pneumonia detection on data from three independent corpuses. The study found that internal model performance was superior to external model performance in all three single dataset comparisons. Where datasets were combined into supersets, the superset trained models outperformed the single dataset trained models – however that performance improvement did not generalize to a third dataset that was not included in the training data supersets. Interestingly, the discussion in this study is focussed on AUC scores declining in generalization tests by 3 – 10 % on the single dataset tests, but does not pay particular attention to the decline in specificity associated with these tests – being in the range of 3 – 34% in single dataset tests and up to 47% on the superset trained models against independent dataset. Sensitivity results for the experiments remained in the 95 – 98% range throughout. Since specificity is a measure of the ability of the model to distinguish disease negative samples from the full sample set, it follows that the poor generalization results from this study flow from high numbers of false positive predictions on the external test set. In other words, the CNN models tested were very poor at separating disease negative sample and tended to classify all samples as disease positive on external testing. This study notes that CNNs may learn so called "confounding factors" which are non-pathological indications of the class to which the sample image belongs, for example sample images from portable scanners are more likely to be disease positive – since these machines are usually used at the bedside of more seriously ill patients (Zech et al., 2018). This study concludes that "Estimates of CNN performance based on test data from hospital systems used for model training may overstate their likely real-world performance."

Given the poor generalization characteristics found by (Zech et al., 2018) from a large number of high quality and consistently acquired images, it is to be expected that the publicly available COVID-19 CT data sets consisting of variable quality images from a large number of sources would exhibit similar or worse generalization problems when used a training data for deep learning models. Moreover, where the disease positive and disease negative classes are independently sourced then the very high likelihood of "confounding factors" caused by systematic differences in source image provenance must be considered, especially in the presence of unrealistically high deep learning classifier accuracy for internal tests. Our experiments seek to confirm whether these systematic biases do indeed exist, and, if so, which image processing techniques are effective at reducing or removing these biases and restoring model generalization.

## 3. Methodology

### 3.1 Dataset selection

A comprehensive survey of COVID-19 data sources was produced by (Shuja et al., 2020) and published in September 2020. This survey documents the following COVID-19 data sets for the CT imaging mode.

The Cohen dataset (J. Cohen et al., 2020) consisting of 125 images with some CT images. Images have been collected from a number of public sources along with an indirect collection from hospitals and physicians. As at the time of writing (28 Feb 2021) this dataset contains only 84 CT images, of which 80 are labelled COVID-19 and 4 labelled as "No Finding". We do not consider this dataset to be of a sufficient size for the purposes of this inter-dataset study. In addition, the CT images contained in this dataset have not been referenced by the associated paper and study (J. P. Cohen, P. Morrison, et al., 2020).

(Zhao et al., 2020) consisting of 812 CT images that have been scraped from the medRxiv (https://medrxiv.org) health sciences, and bioRxiv (https://bioRxiv.org) biology preprint archives. This dataset contains a total of 812 CT images, of which 349 images from 216 patients are labelled COVID-19 and 463 images from only 84 patients are labelled as Non-COVID-19. The dataset is accompanied by an associated study that presents a CNN based deep learning model that achieves accuracy of 89% and an F1 measure of 90%. The size of this dataset and an accompanying deep learning model study makes this a good candidate data set for our purposes and is henceforth referred to as the "COVID-CT" dataset in this paper.

(Wang et al., 2020) consisting of 435 CT images from 99 patients collected both retrospectively and from three associated hospitals located in China. The collected images were a mixture of COVID-19 confirmed cases and other viral pneumonia cases. In this study an Inception based CNN model was trained using transfer learning, achieving an accuracy of 73.1% and an F1 measure of 0.63 against a holdout testing subset. The data from this study is not publicly available and this study has therefore not been included in this paper.

(Shan et al., 2020) used deep learning to develop a COVID-19 infection segmentation/quantification algorithm using a Chinese sourced dataset that is not publicly available. For this reason, this study has not been included in this paper.

(Jun et al., 2020) have assembled a small set of 20 expert segmented and labelled COVID-19 CT scans for the purposes of deep learning based COVID-19 segmentation. This dataset does not include an accompanying deep learning based COVID-19 automated diagnosis study and has therefore not been included in this paper.

Subsequent to the review provided by (Shuja et al., 2020) two further COVID-19 CT studies have become available, both of which contain sizable datasets with an accompanying CNN based deep learning study focussed on COVID-19 diagnosis. Firstly, (Soares et al., 2020) consisting of 2482 CT scans of which 1252 are COVID-19 positive and 1230 are COVID-19 negative from patients who had presented for other pulmonary diseases. Images were collected from hospitals in Sao Paulo, Brazil. The dataset is accompanied by a study that uses a hand-crafted deep learning network achieving an accuracy of 97.38% and an F1 score of 97.31%. The data from this study is publicly available and ideal for our experiments. This dataset is henceforth referred to as the "SARS-COV-2" dataset in this paper. Secondly, (Gunraj et al., 2020) consisting of 194,922 images from 4,501 patient cases initially derived from CT imaging data collected by the China National Center for Bioinformation (CNCB) (K. Zhang et al., 2020), but augmented with images from other data sources. The 2A release of this dataset contains 92,268 COVID-19 positive and 30,749 Normal images with the remainder being labelled as non-COVID-19 pneumonia. The dataset is accompanied by a study that utilizes a hand-crafted CNN based network utilizing skip-connections. From the published confusion matrix, we can calculate that this network achieved an average accuracy of 99.1% with an F1 score of 99.0% across the three classes. The paper notes in discussion that a generalization study would be needed for this model to be considered for clinical use, as well as a better understanding of how the classifier was able to distinguish between COVID-19 pneumonia cases and other pneumonia cases. The COVID-19 and Normal labelled images are useful for comparison with the other datasets described above and have been included in this study. This subset is henceforth referred to as the "COVIDNet-CT-2A" dataset in this paper.

A large dataset sourced from hospitals in Moscow has recently been made available by (Morozov et al., 2020). This dataset contains 1110 CT scans with COVID-19 positive and negative examples. The dataset is unique in that each patient has been assessed by expert radiologists and labelled according to COVID-19 lung involvement stratified at 25% intervals as CT-0 to CT-4 with CT-0 representing 0% lung involvement and CT-4 representing 75-100% lung involvement. This dataset is used as a control in the second experiment of our study to determine the strength of our models ability to predict lung involvement in an independent dataset for which involvement stratification labelling is available. Our hypothesis is that a well generalized model should show an increase in disease positive classification score as the level of lung involvement increases, with average predicted scores for the stratified images increasing in accordance with the strata labels. This dataset is henceforth referred to as the "Moscow" dataset in this paper.

We have introduced one private dataset to this study in order to provide a very controlled corpus with clear provenance from which we can assess factors affecting generalization performance. This dataset comprises 626 COVID-19 positive images from 626 patients, and 619 COVID-19 negative images from 619 patients. These images were extracted from the Mehr Hospital, Iran PACS before being deidentified, with the middle CT slice extracted, compressed into a 512 x 512 PNG image file and zoomed in by 40% to remove the circular scan field mask to make the images structurally similar to the other datasets. Images were originally acquired from a GE Medical Brightspeed 16 detector multislice CT scan machine using a low dose spiral high-resolution CT scan. The source CT scans were independently assessed and matched by two trained and certified radiologists. This dataset is referred to as the "MID-CT" dataset in this paper.

An investigation was undertaken to understand the variety of sources underlying each dataset with the key observation being that the datasets do not always have COVID-19 disease positive and disease negative sample images co-sourced. As discussed in the introduction, this would be highly problematic if these independent source provenances were to impart systematic bias upon the dataset. Such a systematic difference would produce artificially high classification metrics for internal testing, but poor results in external testing. Attributes of the datasets that we would expect to be problematic in this manner include:

1. Non-uniform image size - since resizing images to 244 x 244 pixels to suit classifier input requirements will non-uniformly distort the source images due to differential source image aspect ratios. This issue is notable in the COVID-CT and COVIDNetCT-2A datasets.

2. Inconsistent image resolution – for example images that have been extracted from PDF sources tend to be lower resolution than CT slice extracts from DICOM source files. This issue is notable in the COVID-CT and COVIDNetCT-2A datasets.

3. Inconsistent compression artefacts – where disease positive and disease negative samples come from different sources it is important that compression algorithms match to avoid classifiers learning to separate classes based on systematic differences in compression artefacts. This issue is particularly notable on the COVID-CT dataset.

4. Patients samples with non-COVID-19 pneumonia diagnosis, since the typical radiological findings for COVID-19 pneumonia can be seen in some non-COVID-19 pneumonia (Duzgun et al., 2020). With images downsized to only 244 x 244 pixels it is questionable that these images still contain enough information for a classifier to separate on the subtle differences between these two types of pneumonia. This is a consideration only for the COVIDNetCT-2A dataset. For the purposes of our investigation we have removed the non-COVID-19 pneumonia images from this dataset to support binary classification allowing for comparison with other datasets and models.

5. Inconsistent CT slicing. Some CT slices show very little lung field since due to the image being dominated by the heart as shown in Fig.1 below. These images therefore contain very little, perhaps no COVID-19 disease related pathological information. CT scans represent a moving image locus down the body, and the image properties vary greatly by slice location. A fair training data set would select the same slice, or slices, for each patient for each class to avoid having the classifier learning to separate classes by anatomical features related to slice position rather than pathological finding. In the case of pulmonary disease, the selected slice or slices should maximise the view of lung field, and minimise the obfuscation of the image by the heart, however images such as those shown in Fig. 1 (extracted from the COVIDNet-CT-2A dataset) are very common in the publicly available datasets. This is a feature that is present to some extent in each of the datasets except but is particularly notable in the COVIDNetCT-2A dataset, which includes a variable number of CT slices for some patients, many including such obfuscated views.

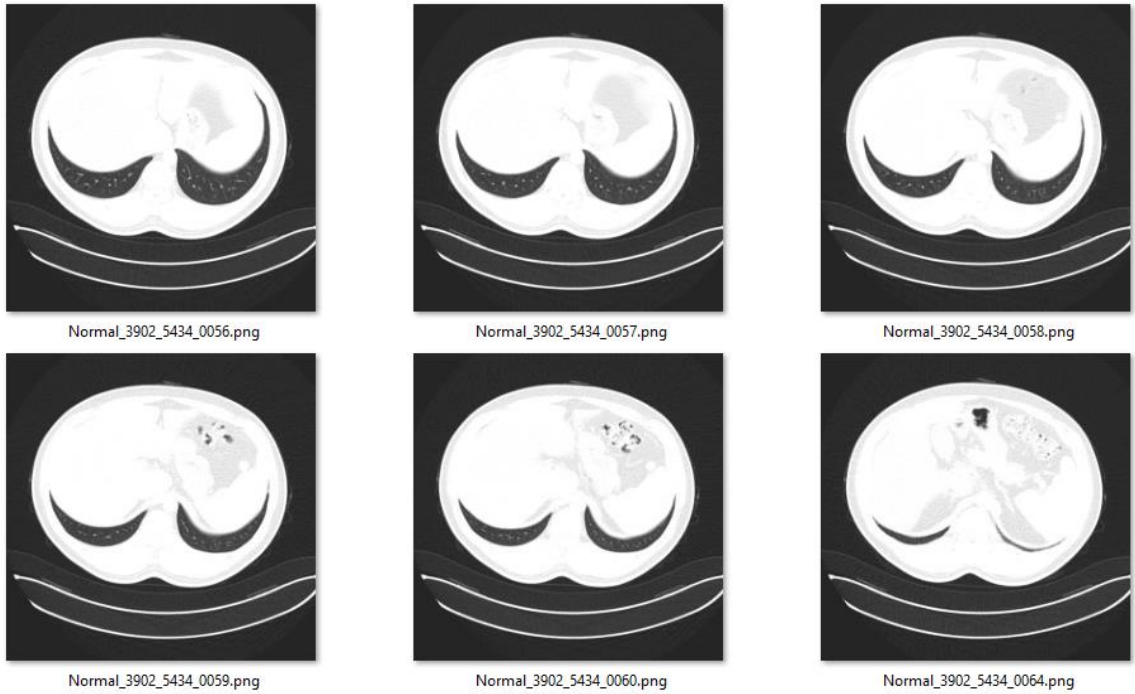

**Fig. 1.** Sample CT images with lung field obfuscated by the heart

It must also be noted that COVIDNet-CT-2A is a superset of images that contains positive samples from the Moscow dataset and negative samples from the COVID-CT dataset, and is therefore initially polluted for the purposes of our generalization study. In assembling the COVIDNet-CT-2A dataset we have removed the COVID-CT and Moscow image samples to remove cross-contamination of the datasets to allow for an unpolluted generalization test. This has the effect that a direct comparison between the internally trained COVIDNet-CT-2A results and ours is invalid although our internal testing results remain very close those of the source study.

Ideally, any systematic variations across datasets of the disease positive and disease negative images would be caused by pathology features only and thereby restricted to brightness differences in the lung field. To test this, we created an average composite image for each dataset and class to provide a visual similarity analysis across the classes. This was achieved by looping through the image files for each class, and recursively joining pairs of images using a weighting function to ensure that every image in the sequence was equally represented in the average composite image. The image frequency histogram for each composite image was calculated and charted in Table 1. Whilst the composite images/histograms for Moscow, MID-CT and SARS-COV-2 datasets are very consistent across classes, both COVID-CT and COVIDNet-CT-2A exhibit significant differences for the different classes. In the case of COVID-CT the corners of the composite image have very different pixel values, with the negative classes being black and the positive classes being a light grey. The image histogram for the COVID-CT disease negative class covers a much broader spectrum than that for the disease positive class. For COVIDNet-CT-2A the average image histograms for the two classes cover a similar frequency spectrum, but the disease positive class amplitude is much lower than that on the disease negative class histogram. From these systematic differences visible between classes in these datasets we hypothesise that deep learning models trained on COVID-CT and COVIDNet-CT-2A are most likely to train on non-pathological features which will limit external generalization.

We also noticed significant differences in zoom level and CT image acquisition artefacts between the datasets. For example, the MID-CT, COVID-CT and SARS-COV-2 datasets are both zoomed to maximise the lung field in the frame. The Moscow images appear within a circular mask and are of a different scale to the other datasets. Since we are using the Moscow images as a control dataset for lung involvement inferencing, we did not attempt to match the zoom level of this dataset to our other datasets preferring to leave this control dataset completely unprocessed. The COVIDNet-CT-2A average composite images contain a series of horizontal arcs in the lower quadrant of the image that are not present in the other image sets. The extent to which these structural differences between the datasets may limit external generalization is considered in the results and discussion sections of this paper.

**Table 1.** Average image composites and observations for selected datasets

| Dataset Reference (chi-squared) | Image Class Average Composites & Histograms | | |
|---|---|---|---|
| | *COVID-19 Negative* | *COVID-19 Positive* | *Observations* |
| **COVID-CT** (7.88) | 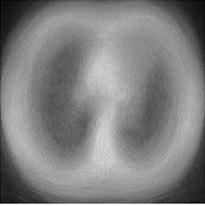 | 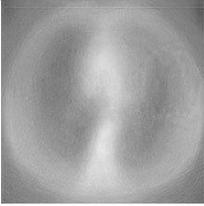 | Multi-sourced from scraped PDF files. Clear differences in brightness and contrast between disease positive and disease negative classes apparent in average composite images. Histogram comparison shows different frequency distributions between classes with disease negative composite showing a much broader frequency spectrum resulting in a relatively high chi-squared value. |
| **COVIDNet-CT-2A** (14.946) | 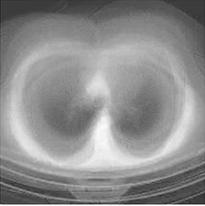 | 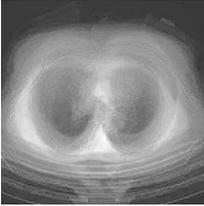 | Multi-sourced from a number of daasets. Some differences in brightness and contrast between disease positive and disease negative classes apparent in average composite images. Histogram comparison different between classes, with the disease negative composite showing higher amplitude across the frequency spectrum resulting in a very high chi-squared value. CT scan artefacts displaying as arcs in lower part of image not present in other datasets. |
| **SARS-COV-2** (7.383) | 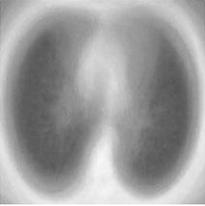 | 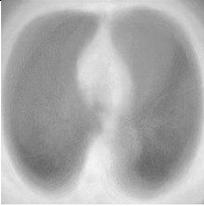 | Sourced from multiple Brazillian clinics but acquired and processed centrally. Brightness and contrast are a systematic match across classes. Histogram comparison relatively similar in amplitide between classes, with disease negative composite showing a slightly broader spectrum. |
| **MID-CT** (1.490) | 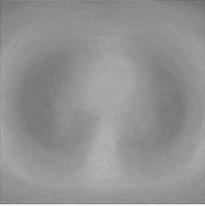 | 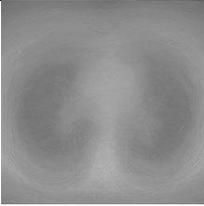 | Single sourced from Iranian Hospital using consistent CT scanning apparatus and acquisition pipeline. Zoomed from 512 x 512 to 350 x 350 and centered to remove circular mask from extracted middle slice. Brightness and contrast systematic match across classes. Histograms almost identical between classes with low chi-squared value. |

| MOSCOW (0.139) | 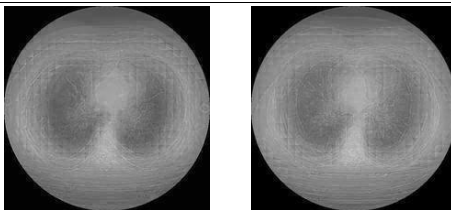 | Sourced from multiple Moscow based clinics but acquired and processed centrally.<br><br>Directly extracted from Source DICOM image with Circular image frame shared only with MID-CT.<br><br>Histogram almost identical between class with very low chi-squared value.<br><br>Compression artefacts are consistent across classes, but reinforced by averaging and clearly visible on composite image. |
|---|---|---|

## 3.2 Pre-processing pipelines

Our study scope covers both the generalization behaviour of trained deep learning models and the effect that commonly used pre-processing techniques have on that generalization behaviour. To meet this objective our models have been trained and tested not only against raw data, but also using simple histogram equalization using parameters matching ImageNet (Krizhevsky et al., 2017) and CLAHE as successfully employed in a number of studies into deep learning based lung pathology detection from medical images (Sarkar et al., 2020; Wajid et al., 2017). We were also interested in the effect of handcrafted feature extraction layers, often implemented as wavelet filters such as Gabor filters, which have been shown to improve the accuracy of deep learning classification networks (Han et al., 2014; Paulraj & Chellliah, 2018; Ye et al., 2007) in thoracic disease imaging applications.

Each combination of image histogram function was trained and tested with and without a learning Gabor filter replacing the first convolution layer in the Densenet-121 network. This allows our experimental work to reveal which, if any, of these techniques have a positive effect on the generalization behavior of the deep learning model under investigation.

## 3.3 Model development

The torchvision (Marcel & Rodriguez, 2010) DenseNet121 (Huang et al., 2017) network architecture was selected for the study which is consistent with other studies successfully using DenseNet121 as a deep learning classifier for COVID-19 image datasets (Pham, 2020).

The DenseNet neural network architecture consists of a sequence of dense blocks comprising stacked convolution layers separated by transition layers composed of a 1 × 1 convolution down-sampling prior to average pooling as input to the next dense block. Each convolution layer in the dense blocks receives as input the feature maps output of all preceding convolution layers in the dense block. This feature map reuse characteristic means that lower levels in the dense block do not need to relearn features from higher levels in the dense block. This architecture is shown in Fig. 2 below.

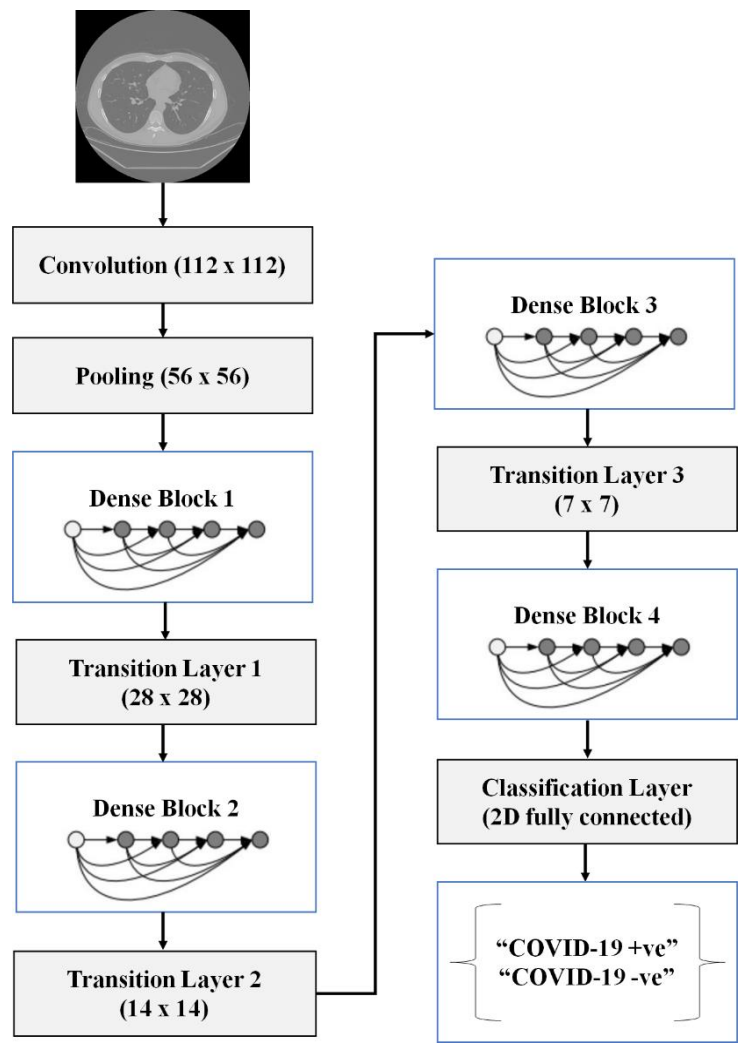

**Fig. 2.** Modified DenseNet-121 CNN used for training and inference

Since our experiments are all binary classifications (COVID-19 disease positive versus COVID-19 disease negative) we replaced the 1000 neuron DenseNet output fully connected layer with 2 neurons as shown in Fig.2 above. Following a number of studies showing the effectiveness of pre-trained network models in thoracic disease classification (Morid et al., 2021), we employed transfer learning to pretrain the DenseNet model in recognising natural shapes from the ImageNet (Pytorch.org, n.d.) training corpus. The entire network was then fine-tuned against the target datasets at a learning rate of 1e-5 using the Adam optimizer (Kingma & Ba, 2017) with standard parameters. Augmentations of 1 degree rotation with expansion and horizontal flip were applied. Class imbalance between disease negative and disease positive classes was addressed using a balancing random sampler. For each model a single training run was executed with the trained model captured for the purposes of external testing and involvement scoring.

To ensure consistent training across the datasets and models, the DenseNet network was trained until three consecutive epochs showed no reduction in validation loss, implementing a regularization technique known as early stopping (Prechelt, 2012). At this point each model was tested against a hold-out test set confirming to the recommended test split of the dataset source for COVID-CT, COVIDNet-2A and SARS-COV-2 and otherwise our selection of 20% of images for the private MID-CT set with no patient overlap between the training and holdout test image samples since this dataset has only a single image per patient. Training history plots for all models are available as supplementary information.

### 3.4 Experiment design

#### 3.4.1 External testing

All combinations of no equalization, simple histogram equalization, application of CLAHE and use of a learning Gabor filter replacing the first convolution layer were scripted, with training completed for all four datasets (reserving the Moscow dataset for the involvement scoring study) resulting in 12 trained models (4 datasets x 3 pre-processing options) without Gabor filtering and 12 trained modes with Gabor filtering. Each of these trained models was then cross validated against the test data sets corresponding to each other model as an external generalization study. This study was performed for each combination of pre-processing option and with/without Gabor filtering as shown in Fig. 3 below.

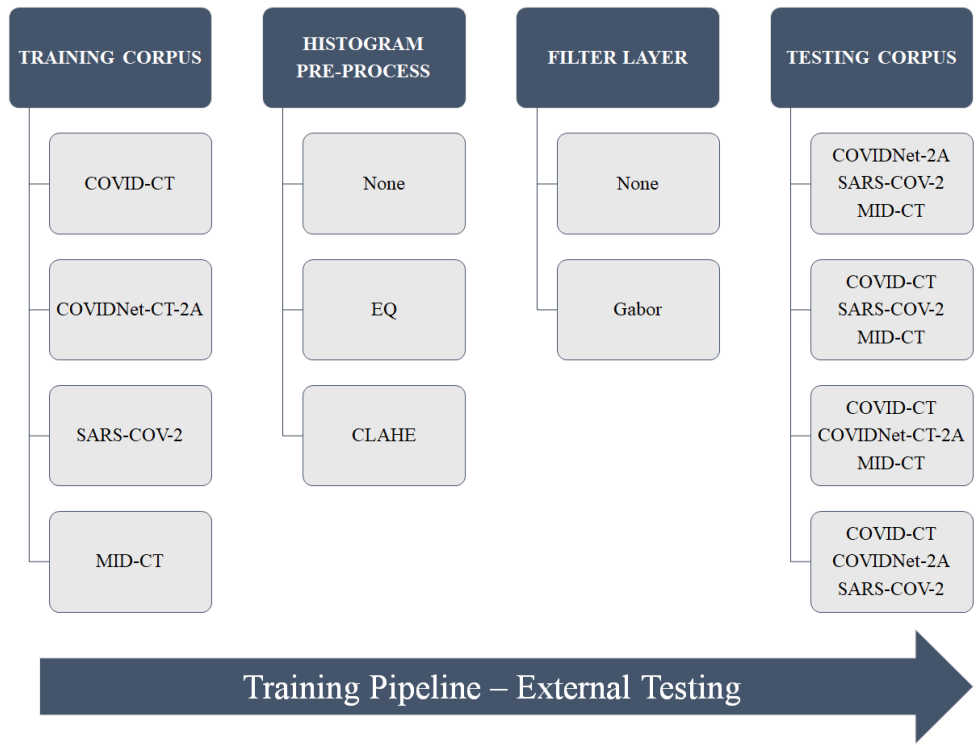

**Fig. 3.** Experimental flow for external testing

*3.4.2 Lung invovelment scoring*

Each trained model was used to score COVID-19 positive and COVID-19 negative lung involvement against each image in the Moscow expertly stratified dataset. For each stratification in the Moscow dataset an average involvement predictive score for each of our models was calculated. The correlation between the average predicted score on the y-axis and involvement strata on the x-axis was plotted for each model to test our hypothesis that a generalized COVID-19 diagnostic model would be predictive of lung involvement strata. Once again, this experiment was performed for each combination of pre-processing option and with/without Gabor filtering as shown in Fig. 4 below.

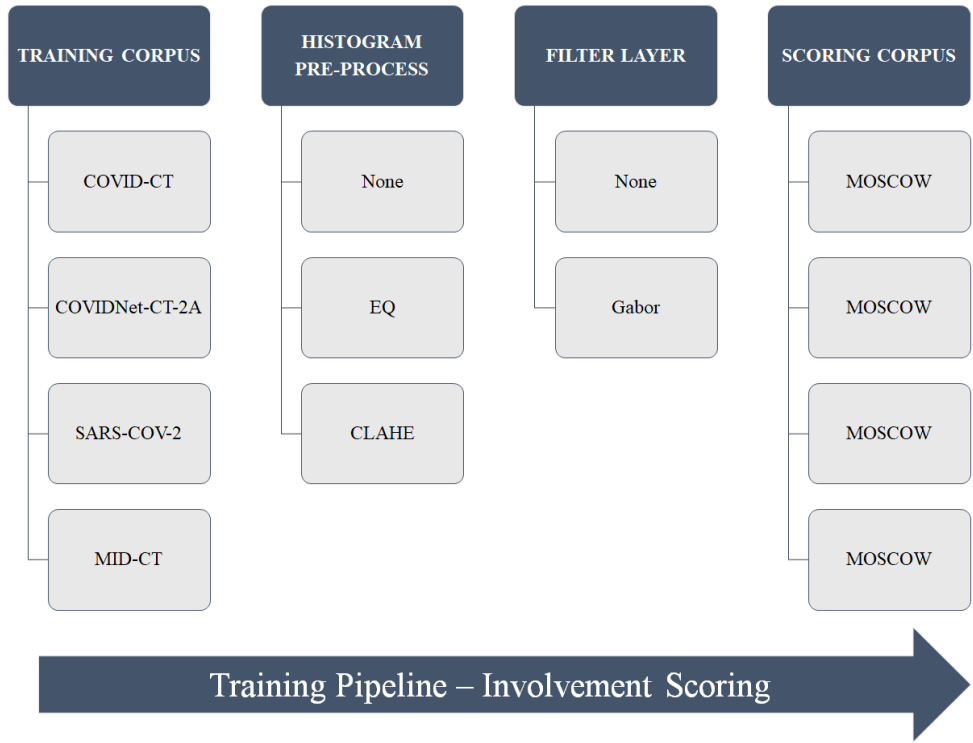

**Fig. 4.** Experimental flow for lung involvement scoring

## 4. Results

*4.1 Internal testing results*

For the purposes of comparison and validation of our DenseNet-121 trained models a summary of results from the source dataset papers is presented in Table 2. Although sensitivity and specificity metrics are usually used in the medical context to assess the effectiveness of predictive models, these were not provided by all of the source studies and our summary is thereby limited to Accuracy, Precision, Recall and calculated F1 score.

**Table 2.** Source dataset results compared

| Study Reference | Source Study Results (%) | | | |
|---|---|---|---|---|
| | *Accuracy* | *F1* | *Precision* | *Recall* |
| COVID-CT | 89.0 | 90.0 | N/A | N/A |
| COVIDNet-CT-2A | 99.1 | 99.0 | 98.8 | 99.2 |
| SARS-COV-2 | 97.4 | 97.3 | 99.1 | 95.5 |

Given the very large number of comparison experiment variations we present our results in a tabular format over Tables 3 - 11 summarizing the results for each dataset and trained model. Each cell in the table presents three numbers, from top to bottom these are to be read in order as the results of no histogram pre-processing, standard histogram equalization pre-processing and CLAHE pre-processing. Separate tables have been created showing the results with and without a learning Gabor filter applied to allow for easier comparison.

Table 3 presents internal testing results for all our models without the application of a Gabor filter replacing the first convolution layer. The first observation is that there is no clear pattern that image pre-processing such as histogram equalization and CLAHE have any significant effect on classification metrics under internal testing with all results very close regardless of pre-processing technique. The only exception to this is the MID-CT models which performed significantly better without either histogram equalization or CLAHE. Noting that this dataset is high quality with consistent brightness and contrast we interpret this as a result of these histogram functions amplifying image noise (Lehr & Capek, 1985) and thereby diminishing contrast of the subtle image opacities associated with COVID-19 pneumonia (Wong et al., 2020) thereby reducing the separability of the classes. The second observation is that our results using an off-the-shelf classifier are comparable to those of the original studies, with a small improvement on the SARS-COV-2 study results from our single training run and slightly lower results against the COVIDNet-CT-2A dataset which may be the result of our removal of the Moscow and COVID-CT images from this set. Our results for the COVID-CT dataset, using the same pipelines, classifier and hyperparameters as used for other datasets is substantially lower than then the original study, whose results we have been previously unable to replicate using off-the-shelf networks (M. Horry et al., 2020).

**Table 3.** Internal results without Gabor filter layer

| Study Reference | Variation | Internal results without learning Gabor filter (%) | | | |
| --- | --- | --- | --- | --- | --- |
| | | *Accuracy* | *F1* | *Precision* | *Recall* |
| COVID-CT | No Preprocess | 69.0 | 69.0 | 73.0 | 69.5 |
| | Histogram Eq | 70.9 | 70.0 | 74.5 | 71.5 |
| | CLAHE | 64.0 | 64.0 | 65.0 | 64.0 |
| COVIDNet-CT-2A | No Preprocess | 98.9 | 98.5 | 98.5 | 99.0 |
| | Histogram Eq | 98.9 | 98.5 | 98.5 | 99.0 |
| | CLAHE | 98.8 | 98.5 | 98.5 | 98.5 |
| SARS-COV-2 | No Preprocess | 97.5 | 97.5 | 97.5 | 97.5 |
| | Histogram Eq | 97.9 | 98.0 | 98.0 | 98.0 |
| | CLAHE | 98.3 | 98.0 | 98.5 | 98.0 |
| MID-CT | No Preprocess | 80.0 | 80.0 | 80.0 | 80.0 |
| | Histogram Eq | 79.0 | 79.0 | 79.0 | 79.0 |
| | CLAHE | 78.0 | 78.0 | 78.0 | 78.0 |

Table 4 presents internal testing results with the application of a learning Gabor filter replacing the first convolution layer. This resulted in improved classification metrics for the COVID-CT, COVIDNet-CT-2A and MID-CT datasets. For COVID-CT, the key improvement in classification metrics observed is improved recall leading to an accuracy gain of 4% when using histogram equalization. This result is still not as good as the original study, but matches our previous best result against this dataset using pre-trained networks (M. Horry et al., 2020). For COVIDNet-CT-2A we found that using a combination of Gabor filtering and CLAHE pushed our internal accuracy for COVIDNet-CT-2A to 99.1% which is equivalent to the original study, bearing in mind that we have removed the COVID-CT and Moscow data sets from our version of COVIDNet-CT-2A in order to avoid polluting our external generalization study. For MID-CT we found that the combination of Gabor filtering and CLAHE resulted in an internal accuracy of 86% which is our best internal score for this dataset by a significant margin of 6%.

**Table 4.** Internal results with Gabor filter layer

| Study Reference | Variation | Internal Results with Learning Gabor Filter (%) | | | |
|---|---|---|---|---|---|
| | | *Accuracy* | *F1* | *Precision* | *Recall* |
| COVID-CT | No Preprocess | 63.5 | 62.5 | 96.0 | 64.5 |
| | Histogram Eq | 74.9 | 75.0 | 76.0 | 75.5 |
| | CLAHE | 70.0 | 70.0 | 71.5 | 70.5 |
| COVIDNet-CT-2A | No Preprocess | 98.6 | 98.5 | 98.5 | 98.5 |
| | Histogram Eq | 99.0 | 98.5 | 98.5 | 99.5 |
| | CLAHE | 99.1 | 99.0 | 98.5 | 99.5 |
| SARS-COV-2 | No Preprocess | 97.9 | 98.0 | 98.0 | 98.0 |
| | Histogram Eq | 96.3 | 96.0 | 96.0 | 96.0 |
| | CLAHE | 95.8 | 96.0 | 96.0 | 96.0 |
| MID-CT | No Preprocess | 79.0 | 79.0 | 79.0 | 79.0 |
| | Histogram Eq | 79.0 | 79.0 | 79.0 | 79.0 |
| | CLAHE | 86.0 | 86.0 | 86.0 | 86.0 |

These internal testing results show that it is possible to achieve competitive results to the published studies using off-the-shelf CNNs with transfer learning with the exception of the COVID-CT dataset, being a challenging dataset for several reasons previously noted and which remains unpublished at the time of writing.

*4.2   External testing results*

Having trained each model for each combination of pre-processing technique with and without a learning Gabor filter replacing the first convolution layer we proceeded to externally test each model against each other training corpus test data partition.

*4.2.1   External testing of COVIDNet-CT-2A models*

Tables 5 and 6 show the results of external testing of the COVIDNet-CT-2A trained models.

**Table 5.** COVIDNet-CT-2A external results without Gabor filter layer

| Model Dataset: COVIDNet-CT-2A | Internal results without learning Gabor filter (%) | | | | | |
|---|---|---|---|---|---|---|
| | Variation | Accuracy | F1 | Precision | Recall | Sensitivity | Specificity |
| COVID-CT | No Preprocess | 38.7 | 37.0 | 50.0 | 50.0 | 83.1 | 16.9 |
| | Histogram Eq | 40.0 | 38.5 | 51.0 | 51.0 | 83.3 | 18.7 |
| | CLAHE | 62.3 | 61.0 | 62.0 | 63.5 | 67.5 | 59.8 |
| MID-CT | No Preprocess | 50.0 | 35.0 | 50.0 | 50.0 | 98.0 | 2.0 |
| | Histogram Eq | 53.0 | 42.5 | 61.5 | 53.0 | 96.0 | 10.0 |
| | CLAHE | 59.0 | 57.0 | 61.5 | 59.0 | 82.0 | 36.0 |
| SARS-COV-2 | No Preprocess | 75.7 | 75.5 | 77.0 | 80.5 | 93.8 | 66.8 |
| | Histogram Eq | 60.9 | 60.5 | 72.0 | 70.5 | 98.2 | 42.5 |
| | CLAHE | 63.0 | 61.0 | 72.5 | 72.0 | 98.2 | 45.6 |

**Table 6.** COVIDNet-CT-2A external results with Gabor filter layer

| Model Dataset: COVIDNet-CT-2A | Internal results with learning Gabor filter (%) | | | | | |
|---|---|---|---|---|---|---|
| | Variation | Accuracy | F1 | Precision | Recall | Sensitivity | Specificity |
| COVID-CT | No Preprocess | 60.2 | 60.0 | 63.0 | 64.5 | 77.1 | 51.9 |
| | Histogram Eq | 54.1 | 54.0 | 58.0 | 54.0 | 73.1 | 44.9 |
| | CLAHE | 44.7 | 44.5 | 52.5 | 52.0 | 74.5 | 30.1 |
| MID-CT | No Preprocess | 50.2 | 50.0 | 60.5 | 59.5 | 85.9 | 32.6 |
| | Histogram Eq | 60.9 | 61.0 | 70.5 | 69.5 | 95.2 | 44.0 |
| | CLAHE | 69.6 | 69.5 | 73.0 | 75.5 | 91.9 | 58.7 |
| SARS-COV-2 | No Preprocess | 77.9 | 70.0 | 80.5 | 68.5 | 40.7 | 96.2 |
| | Histogram Eq | 84.4 | 83.0 | 82.0 | 84.0 | 82.7 | 85.3 |
| | CLAHE | 77.9 | 73.0 | 76.0 | 71.5 | 52.6 | 90.4 |

The COVIDNet-CT-2A trained models did not generalize well to either the COVID-CT or MID-CT datasets regardless of pre-processing histogram function or presence of a learning Gabor filter. Our earlier investigation into the average composite image of the COVIDNet-CT-2A dataset showed that this dataset is, on average, structurally different due to closer zooming in COVID-CT and MID-CT . External testing against the SARS-COV-2 dataset without Gabor filter was good with best accuracy of 75.7% and very good with a learning Gabor filter with a best accuracy of 84.4%.

*4.2.2 External testing of COVID-CT models*

Tables 7 and 8 show the results of external testing of the COVID-CT trained models. The challenging nature of this dataset has been previously described and our expectation that these models would not generalize is supported by the external testing results with classification metrics that show no significant class separation regardless or pre-processing option or presence of a learning Gabor filter.

We interpret this poor generalization performance as evidence that systematic differences between the COVID-19 disease positive and COVID-19 disease negative image sets, resulting from different data source provenances, have led to the classifier learning to separate classes using non-pathological features. This training has not translated to other datasets not sharing these non-pathological features.

Table 7. COVID-CT external results without Gabor filter layer

| Model Dataset: COVID-CT | Variation | Internal results without learning Gabor filter (%) | | | | | |
|---|---|---|---|---|---|---|---|
| | | Accuracy | F1 | Precision | Recall | Sensitivity | Specificity |
| COVIDNet-CT-2A | No Preprocess | 51.2 | 40.5 | 62.5 | 53.0 | 96.9 | 8.6 |
| | Histogram Eq | 55.7 | 49.5 | 65.0 | 57.0 | 93.9 | 9.4 |
| | CLAHE | 53.2 | 45.0 | 63.0 | 54.5 | 94.9 | 14.3 |
| MID-CT | No Preprocess | 50.2 | 41.5 | 54.5 | 51.5 | 91.8 | 11.4 |
| | Histogram Eq | 47.8 | 34.0 | 44.0 | 49.5 | 96.9 | 1.9 |
| | CLAHE | 48.3 | 32.5 | 24.0 | 50.0 | 100.0 | 0.0 |
| SARS-COV-2 | No Preprocess | 52.7 | 52.5 | 53.0 | 53.0 | 56.1 | 49.5 |
| | Histogram Eq | 43.3 | 41.0 | 42.5 | 44.0 | 66.3 | 21.9 |
| | CLAHE | 47.8 | 47.5 | 48.0 | 48.0 | 58.1 | 38.1 |

Table 8. COVID-CT external results with Gabor filter layer

| Model Dataset: COVID-CT | Variation | Internal results with learning Gabor filter (%) | | | | | |
|---|---|---|---|---|---|---|---|
| | | Accuracy | F1 | Precision | Recall | Sensitivity | Specificity |
| COVIDNet-CT-2A | No Preprocess | 51.2 | 43.5 | 56.0 | 52.5 | 90.8 | 14.3 |
| | Histogram Eq | 51.2 | 42.5 | 58.5 | 52.5 | 92.9 | 11.4 |
| | CLAHE | 51.7 | 48.5 | 54.0 | 53.0 | 80.6 | 24.8 |
| MID-CT | No Preprocess | 53.7 | 43.5 | 75.5 | 55.0 | 100.0 | 10.5 |
| | Histogram Eq | 52.7 | 42.0 | 71.0 | 54.5 | 99.0 | 9.5 |
| | CLAHE | 47.8 | 32.5 | 24.0 | 49.5 | 99.0 | 0.0 |
| SARS-COV-2 | No Preprocess | 48.3 | 47.5 | 48.0 | 48.0 | 36.7 | 59.0 |
| | Histogram Eq | 45.8 | 45.0 | 46.0 | 46.0 | 62.2 | 30.5 |
| | CLAHE | 49.8 | 49.5 | 50.0 | 50.0 | 54.1 | 45.7 |

*4.2.3 External testing of MID-CT models*

Tables 9 and 10 show the results of external testing of the privately acquired MID-CT trained models. Despite the uniformity and high quality of this dataset, the models based on this dataset have failed to generalize to any other dataset. We interpret this as a result of single slice selection (the middle slice) providing too little image variation in comparison to the other datasets which have CT slices taken from a large number of locations on the patient. Since the Moscow dataset is also a single slice per patient dataset, we expect improved results on lung involvement scoring against the Moscow dataset in our second experiment.

**Table 9.** MID-CT external results without Gabor filter layer

| Model Dataset: MID-CT | | Internal results without learning Gabor filter (%) | | | | | |
|---|---|---|---|---|---|---|---|
| | Variation | Accuracy | F1 | Precision | Recall | Sensitivity | Specificity |
| COVID-CT | No Preprocess | 50.0 | 33.5 | 25.0 | 50.0 | 0.0 | 100.0 |
| | Histogram Eq | 50.0 | 33.5 | 25.0 | 50.0 | 0.0 | 100.0 |
| | CLAHE | 50.0 | 33.5 | 25.0 | 50.0 | 0.0 | 100.0 |
| COVIDNet-CT-2A | No Preprocess | 50.0 | 35.0 | 50.0 | 50.0 | 98.0 | 2.0 |
| | Histogram Eq | 53.0 | 42.5 | 61.5 | 53.0 | 96.0 | 10.0 |
| | CLAHE | 59.0 | 57.0 | 61.5 | 59.0 | 82.0 | 36.0 |
| SARS-COV-2 | No Preprocess | 53.0 | 45.5 | 57.0 | 53.0 | 90.0 | 16.0 |
| | Histogram Eq | 50.0 | 33.5 | 25.0 | 50.0 | 100.0 | 0.0 |
| | CLAHE | 53.0 | 51.0 | 53.5 | 53.0 | 32.0 | 74.0 |

**Table 10.** MID-CT external results with Gabor filter layer

| Model Dataset: MID-CT | | Internal results with learning Gabor filter (%) | | | | | |
|---|---|---|---|---|---|---|---|
| | Variation | Accuracy | F1 | Precision | Recall | Sensitivity | Specificity |
| COVID-CT | No Preprocess | 50.0 | 33.5 | 25.0 | 50.0 | 0.0 | 100.0 |
| | Histogram Eq | 50.0 | 33.5 | 25.0 | 50.0 | 0.0 | 100.0 |
| | CLAHE | 50.0 | 33.5 | 25.0 | 50.0 | 0.0 | 100.0 |
| COVIDNet-CT-2A | No Preprocess | 51.0 | 35.5 | 75.5 | 51.0 | 100.0 | 2.0 |
| | Histogram Eq | 50.0 | 33.5 | 25.0 | 50.0 | 100.0 | 0.0 |
| | CLAHE | 50.0 | 38.0 | 50.0 | 50.0 | 94.0 | 6.0 |
| SARS-COV-2 | No Preprocess | 50.0 | 33.5 | 25.0 | 50.0 | 0.0 | 100.0 |
| | Histogram Eq | 68.0 | 66.5 | 72.0 | 68.0 | 46.0 | 90.0 |
| | CLAHE | 67.0 | 63.5 | 77.5 | 67.0 | 36.0 | 98.0 |

*4.2.4 External testing of SARS-COV-2 models*

Tables 11 and 12 show the results of external testing of the SARS-COV-2 trained models. Although performance of these models against the COVID-CT and MID-CT datasets was very poor (with a strong tendency to classify the majority of samples as disease positive) generalization of these models against the COVIDNet-CT-2A was very good with best performance from the histogram equalized model achieving an F1 score of 86% along with 85% and 86.7% from sensitivity and specificity respectively. The improves upon the good performance of the COVIDNet-CT-2A model against the SARS-COV-2 data previously described, and in contrast to those results a Gabor filter was not needed. We interpret these findings as evidence that a smaller, high quality dataset (SARS-COV-2 with 812 high quality images) has produced a more generalizable model than a larger diversified dataset (COVIDNet-CT-2A with 163,630 images) with sample provenance from multiple sources.

**Table 11.** SARS-COV-2 external results without Gabor filter layer

| Model Dataset: SARS-COV-2 | Variation | Internal results without learning Gabor filter (%) | | | | | |
|---|---|---|---|---|---|---|---|
| | | *Accuracy* | *F1* | *Precision* | *Recall* | *Sensitivity* | *Specificity* |
| COVID-CT | No Preprocess | 50.0 | 34.5 | 50.0 | 50.0 | 98.3 | 1.7 |
| | Histogram Eq | 51.3 | 37.5 | 61.0 | 51.0 | 98.3 | 4.2 |
| | CLAHE | 50.4 | 37.0 | 53.0 | 50.5 | 96.7 | 4.2 |
| COVIDNet-CT-2A | No Preprocess | 82.9 | 83.0 | 83.0 | 82.5 | 83.3 | 82.5 |
| | Histogram Eq | 85.8 | 86.0 | 85.5 | 86.0 | 85.0 | 86.7 |
| | CLAHE | 83.8 | 84.0 | 84.0 | 83.5 | 89.2 | 78.3 |
| MID-CT | No Preprocess | 61.3 | 60.5 | 62.5 | 61.5 | 75.8 | 46.7 |
| | Histogram Eq | 48.8 | 38.5 | 46.0 | 48.5 | 90.0 | 7.5 |
| | CLAHE | 51.7 | 39.0 | 60.5 | 51.5 | 97.5 | 5.8 |

**Table 12.** SARS-COV-2 external results with Gabor filter layer

| Model Dataset: SARS-COV-2 | Variation | Internal results with learning Gabor filter (%) | | | | | |
|---|---|---|---|---|---|---|---|
| | | *Accuracy* | *F1* | *Precision* | *Recall* | *Sensitivity* | *Specificity* |
| COVID-CT | No Preprocess | 50.4 | 35.0 | 58.5 | 50.5 | 99.2 | 1.7 |
| | Histogram Eq | 51.3 | 37.5 | 61.0 | 51.0 | 98.3 | 4.2 |
| | CLAHE | 50.0 | 36.0 | 47.0 | 49.5 | 95.8 | 3.3 |
| COVIDNet-CT-2A | No Preprocess | 81.3 | 81.0 | 81.5 | 81.5 | 81.7 | 80.8 |
| | Histogram Eq | 85.0 | 85.0 | 85.0 | 85.0 | 83.3 | 86.7 |
| | CLAHE | 84.6 | 84.5 | 84.5 | 84.5 | 88.3 | 80.8 |
| MID-CT | No Preprocess | 64.2 | 63.5 | 64.5 | 64.5 | 75.8 | 52.5 |
| | Histogram Eq | 50.0 | 41.0 | 50.0 | 50.0 | 89.2 | 10.8 |
| | CLAHE | 50.8 | 37.0 | 56.0 | 50.5 | 97.5 | 4.2 |

## 4.3 Lung involvement scoring results

Having shown some success in external generalization testing we proceeded to test each trained model as an inferencing classifier against the Moscow data set which had been held out as a control. The Moscow dataset is unique in the publicly available COVID-19 CT datasets in that it is a high-quality dataset that has been labelled by expert radiologists with a lung involvement score from 0 to 100% as defined in Table 13.

**Table 13.** Lung involvement strata definitions

| Stratification | Lung involvement (%) |
|---|---|
| CT-0 | 0 |
| CT-1 | 0 – 25 |
| CT-2 | 25 – 50 |
| CT-3 | 50 – 75 |
| CT-4 | 75 – 100 |

Since COVID-19 pneumonia is indicated on CT imagery as lung involvement(Kwee & Kwee, 2020) it follows that generalization of our trained models against the Moscow data set would be evidenced by an increasing probably of COVID-19 disease positive score corresponding to an increasing lung involvement strata. To test this hypothesis, we used each trained model, with each pre-processing option and with/without learning Gabor filters, to score each of the Moscow dataset images for COVID-19 probability. Average predicted score for each stratification was plotted on the y-axis against which we plot ground truth lung involvement stratifications on the x-axis. The results of these experiments are presented in Figs. 5 – 10 below. For models with lung involvement predictive value we expect a line representing COVID-19 disease positive scores to have a positive gradient as involvement score increases from CT-0 to CT-4, and conversely the line representing COVID-19 disease negative scores should have a negative gradient. These gradients should be more pronounced the better a model has performed against the Moscow data set.

Figs. 5(a) and (b) below represent results from trained models without histogram equalization or CLAHE pre-processing, with Fig. 5(b) showing results of adding a learning Gabor filter in place of the first convolution layer of the network. The expected pattern for lung involvement prediction is evident for both the SARS-COV-2 and MID-CT models, although the performance of the MID-CT based model is considered better as it is more effective than the SARS-COV-2 based model in predicting the normal condition represented by data points at CT-0. Fig. 5(b) shows that addition of a Gabor filter had the effect of reducing the performance of both MID-CT and SARS-COV-2 models.

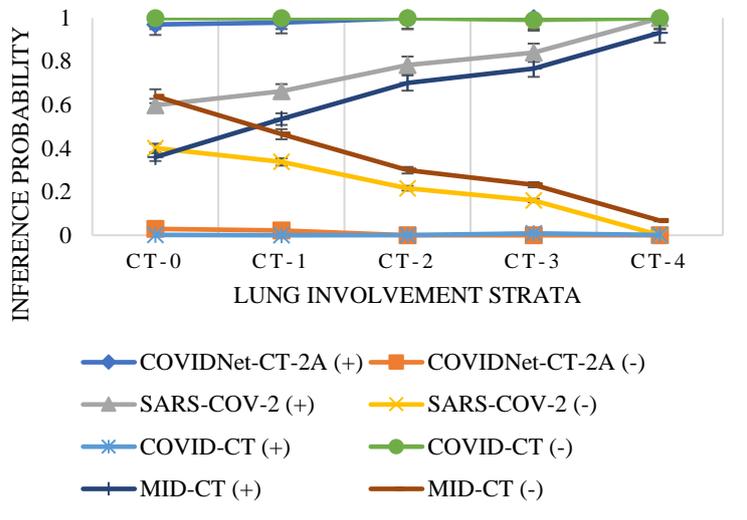 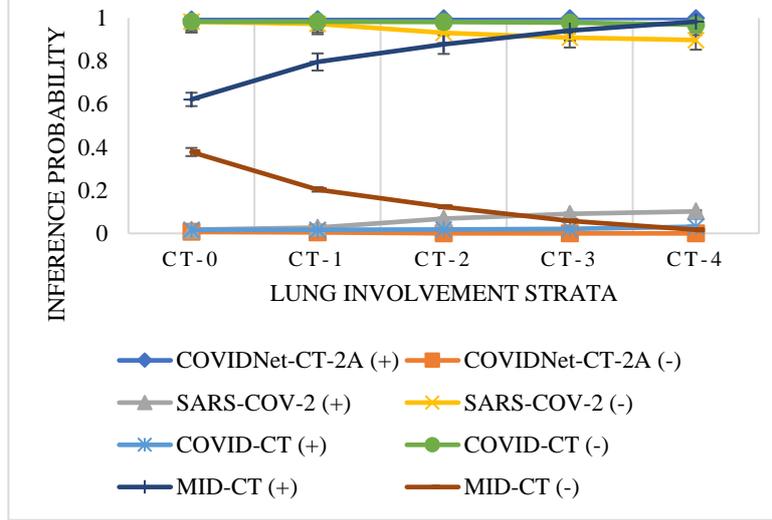

(a). Without Gabor filter layer            (b). With Gabor filter layer

**Fig. 5.** Lung involvement scoring with no histogram pre-processing (95% confidence)

Figs. 6(a) and (b) represent results from trained models with standard histogram equalization. Fig. 6(b) shows the effect of a learning Gabor filter in the position of the first convolution layer of the network. Histogram equalization has eroded the good generalization performance of MID-CT and SARS-COV-2 based models without histogram equalization shown at Fig. 5(a). Application of a Gabor filter to the models further eroded these results as shown in Fig. 6(b). From this, one could conclude that standard histogram equalization alone, and the combination of standard histogram equalization and learning Gabor filter is not an effective combination of techniques to promote generalization.

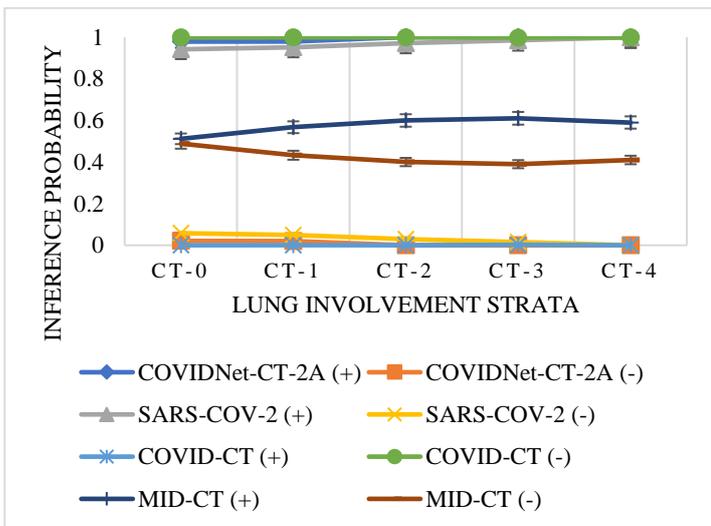 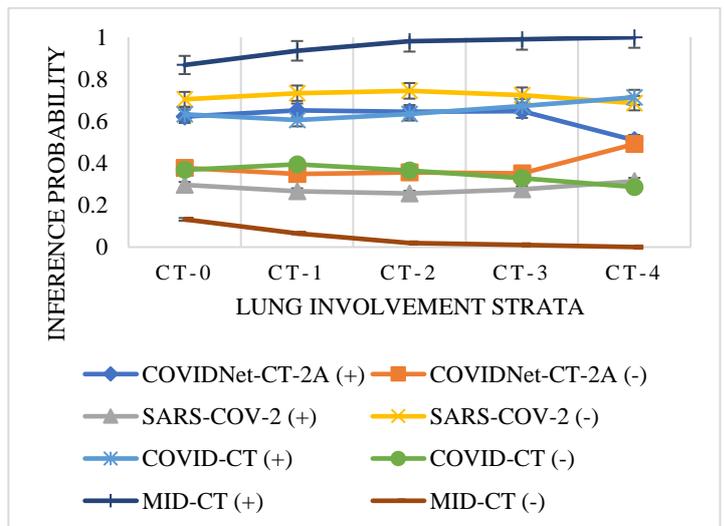

(a). Without Gabor filter layer            (b). With Gabor filter layer

**Fig.6.** Lung involvement scoring with histogram equalization pre-processing (95% confidence)

Figs. 7(a) and (b) represent results from trained models with CLAHE image processing. Fig. 7(b) shows the effect of CLAHE image processing with a learning Gabor filter in the position of the first convolution layer of the network. CLAHE has enhanced the good generalization performance of SARS-COV-2 based model over no pre-processing, and allowed the COVIDNet-CT-2A to show minimal generalization where none was previously evident in Fig.6(a), but eroded the generalization performance of MID-CT. The SARS-COV-2 based model is now able to correctly predict a low probability for disease positive at CT-0 and a high probability for disease positive at CT-4 with a linear gradient. With the Gabor filter included the models, we achieved better results only for SARS-COV-2 based model at the lower stratifications of lung involvement (CT-0 and CT-1) as shown by Fig.7(b).

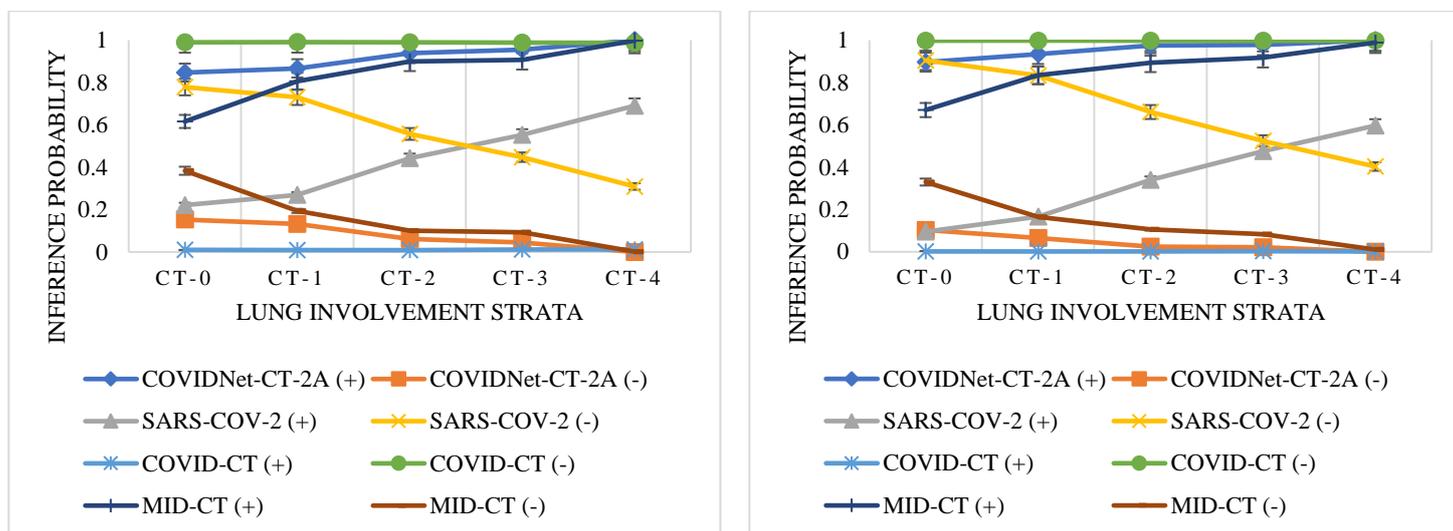

(a). Without Gabor filter layer　　　　　　　　　　(b). With Gabor filter layer

**Fig.7.** Lung involvement scoring with CLAHE pre-processing (95% confidence)

Of our models tested, those based on SARS-COV-2 exhibited superior generalization performance over the other models tested, which led us to focus on which combination of techniques would maximise the predictive value of these models against the Moscow dataset. Inspection of the results graphs for the SARS-COV-2 models showed that application of CLAHE provided the best results, with a Gabor filter providing highest accuracy at CT-0 (minimum lung involvement) representing an improvement in sensitivity and absence of the Gabor filter providing highest accuracy at CT-4 (maximum lung involvement) representing an improvement in specificity.

We combined these SARS-COV-2 based models to use CLAHE as a pre-process and the minimum predicted probability from the Gabor/non-Gabor models as the COVID-19 disease negative score and the maximum probability from the Gabor/non-Gabor models as the COVID-19 disease positive score. Fig. 8 below shows the results of this min-max Gabor ensemble. The ensemble results from this model are excellent, achieving an average score of 75% for the Moscow images without lung involvement and an average score of 96% for the Moscow images with 75-100% lung involvement, with an approximately continuous gradient for the stratifications between these extremes. The SARS-COV-2 based models have generalized well to the Moscow dataset when using this ensembling technique providing very accurate predictive scoring of lung involvement. Table 14 provides the average predictions by involvement stratification for this ensemble model.

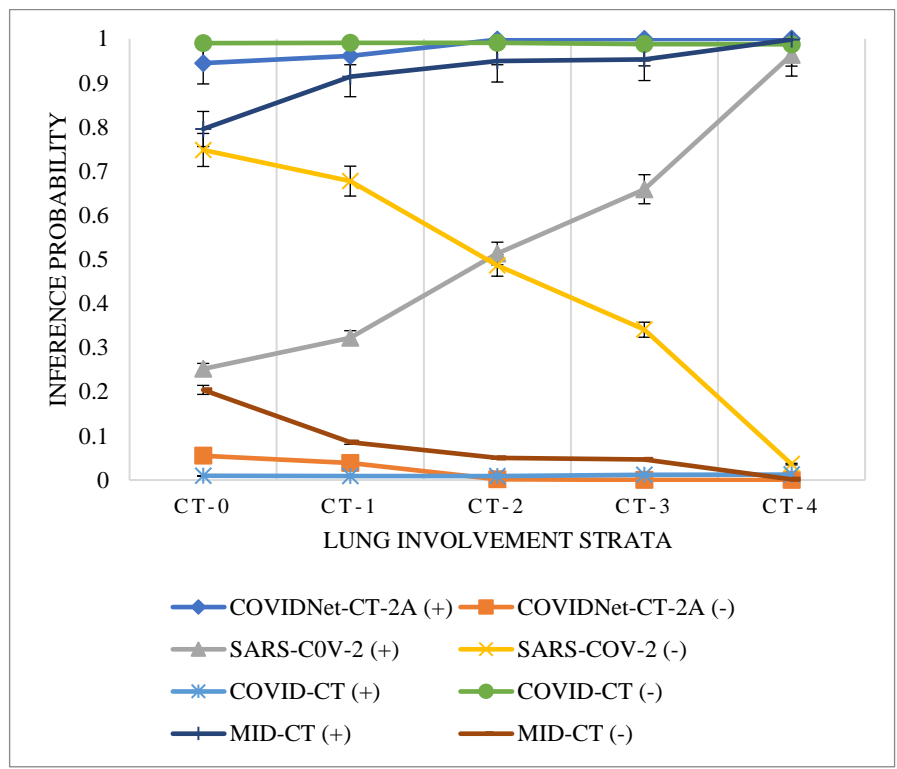

**Fig.8.** Lung involvement scoring with CLAHE and Min-Max Gabor ensemble (95% confidence)

**Table 14.** SARS-COV-2 CLAHE model with Min-Max Gabor ensemble average prediction per strata

| Involvement Strata | Ground Truth Involvement % | Average Prediction SARS-C0V-2 (+) % | Average Prediction SARS-COV-2 (-) % |
|---|---|---|---|
| CT-0 | 0 | 25.184 | 74.816 |
| CT-1 | 0 – 25 | 32.240 | 67.760 |
| CT-2 | 25 – 50 | 51.353 | 48.647 |
| CT-3 | 50 – 75 | 65.917 | 34.083 |
| CT-4 | 75 - 100 | 96.380 | 3.620 |

# 5. Discussion

## 5.1 *Effect of differential image class provenance and patient image diversity*

Our experiments show that attributes of the training datasets had a significant impact on the predictive capability of generated models against external datasets. Our poorest performing models were models based on the COVID-CT dataset. These models did not generalize in either external testing or involvement scoring against the Moscow dataset. We found in internal testing that both histogram equalization and Gabor filter had a positive effect on metrics for this dataset, with these techniques combined achieving a best f1 score of 75%. We interpret the positive effect of histogram equalization on these results to a reduction in the variability of brightness and contrast amongst source images that was very evident in this dataset. Applying a Gabor filter would have the effect of reducing dimensionality manifesting in these experiments as an improvement in specificity since the Gabor filter reinforces pneumonia related features that would be absent from disease negative images. However, since models based on the COVID-CT dataset did not generalize to any external datasets we must ask what attributes of this dataset negatively impacted generalization. We previously noted that this dataset has different sources for disease positive and disease negative images and interpret the observed poor generalization of models trained on COVID-CT to the classifier learning on systematic differences between these classes that are related to systematically different sample image provenance rather than pathological features. Whilst the combination of histogram equalization and Gabor filter reinforced features allowing for improved the internal class separability, these features were likely "confounding factors" that would not be features of external datasets, resulting in poor external generalization. Additionally, we noted previously that the disease positive images for this dataset are sourced from 216 patients, whilst the disease negative images are sourced from only 84 patients. This very low number of disease negative patients would be insufficient for a machine learning classifier to learn generalized disease pathology features. We suspect that the COVID-CT model's ability to detect pathology is dwarfed by such "confounding factors" resulting in very poor performance in our generalization studies. Given these considerations it is not surprising that models based on this dataset were not predictive of lung involvement against the Moscow stratified dataset.

*5.2 Effect of single vs multiple CT slice locations in corpus*

The performance of the MID-CT dataset-based models in external testing was very poor, with the only tests improving up blind chance doing so at the expense of sensitivity or specificity, i.e. a tendency to classify the vast majority of samples as either disease positive or disease negative regardless of ground truth. The MID-CT dataset did have disease positive and disease negative classes co-sourced and subjected to a consistent acquisition pipeline, and furthermore the number of patients in disease positive and disease negative classes was relatively balanced. However, the MID-CT dataset only has a single CT image slice per patient taken from a consistent location, which is different to the other datasets used in external testing, which all have image slices taken from multiple CT scan locations. We attribute the poor generalization performance of MID-CT based models to this mismatch on CT scan image location diversity. Despite the poor results in external testing, models based on the MID-CT dataset did in-fact prove to be predictive for lung involvement score against the Moscow stratified dataset. The Moscow dataset is also a composed of a single scan image per patient and therefore matches MID-CT in terms of patient location image diversity, resulting in the reasonable accuracy in lung involvement scoring for the unprocessed model based on the MID-CT dataset.

*5.3 Comparison of large diverse multi-sourced image corpus vs smaller but consistent image corpus*

The COVIDNet-CT-2A based models showed good generalization in external testing against the SARS-COV-2 dataset. In contrast, under involvement testing against the Moscow dataset, the COVIDNet-CT-2A models had little to no predictive value using no pre-processing or simple histogram equalization and only minimal predictive value with CLAHE pre-processing. This contrasts with the SARS-COV-2 dataset-based models which all showed good generalization against the Moscow dataset, with particularly good results when CLAHE pre-processing was used. These results provide evidence that a smaller, higher quality dataset with disease positive and negative samples co-sourced and consistently acquired (SARS-COV-2) is more effective in generalization to external datasets than a very large, more diverse, dataset composed from multiple sources with a variety of acquisition pipelines (COVIDNet-CT-2A).

*5.4     Effect of image pre-processing and learning Gabor filter*

We found that image histogram equalization and CLAHE and Gabor filtering had a negative effect on MID-CT based model external classification results despite the combination of CLAHE and Gabor filtering providing significantly better results in internal testing for models based on this dataset. We conclude that image pre-processing and Gabor filtering have most likely reinforced signal noise in the manner noted by [63], rather than reinforcing pathological signal only. The MID-CT dataset was single sourced from with fixed CT apparatus operating parameters and therefore very consistent in terms of brightness and contrast. We propose that image histogram pre-processing and wavelet filtering techniques such as Gabor filtering are of use only where differences in image properties such as brightness and contrast have been imparted on the data corpus by different CT apparatus or image acquisition pipelines, as is the case with the SARS-COV-2 dataset which was sourced from multiple Brazil located clinics and shows moderate variability in terms of image size, brightness and contrast. For the SARS-COV-2 we found that histogram equalization and CLAHE improved internal testing results where a Gabor filter was not employed, and also had a positive effect on external testing against the COVIDNet-CT-2A dataset both with and without Gabor filtering.

Finally, we noted that the use of a Gabor filter following CLAHE pre-processing improved the specificity of the SARS-COV-2 based models (i.e. improved classification of the disease negative images), whilst the model with CLAHE but without the Gabor filter showed superior sensitivity. By combining these models with a Min-Max algorithm we produced an ensemble model that accurately predicted lung involvement ground truth against the Moscow dataset being a completely independent dataset. The accuracy of this combined model is remarkable when one considers that the SARS-COV-2 and Moscow datasets have been sourced from the different geographical regions, and that these datasets show significant structural differences on their average image composite.

*5.5     Key principles promoting model generalization*

From these experiments we can suggest the following key principles to be adhered to in assembling medical image corpuses for the purposes of training deep learning medical image computer vision systems where model generalization is an objective:

a. Disease positive and Disease negative classes should be consistently (though not necessarily single) sourced and subjected to consistent image acquisition/compression pipelines. This will minimize systematic structural differences between the classes and eliminate sampling bias resulting in model training on non-pathological "confounding factors". Our SARS-COV-2 and MID-CT models met these criteria and generalized well externally even though these dataset and the Moscow images set were structurally quite different, suggesting that consistency in the model training data is more important that structural consistency across training and external inferencing datasets.

b. Image histogram equalization techniques are useful for improving internal and external generalization results where there is moderate variability in image corpus brightness and contrast (SARS-COV-2) due to multiple sourcing and image acquisition pipelines. These techniques are ineffective where the image set is very consistent (MID-CT) or where there are major inconsistencies in image attributes across the datasets (COVID-CT and COVIDNet-CT-2A).

c. The diversity of images and patients in disease positive classes should be a close match to the those for disease negative classes. This will ensure that that equivalent variation exists in the source data corpuses to ensure that deep learning models do not train on patient similarity rather than pathological features. A diverse selection of patient CT slice positions in the dataset serves to augment patient/image diversity and help models to generalize. Evidence for this proposition comes from the much better lung involvement scoring results achieved from the SARS-COV-2 based models trained on multiple slices per patient compared with the MID-CT based models trained on a single slice per patient.

## 6. Conclusion

Computer vision based medical diagnostic and severity scoring systems cannot be considered be clinically mature until they are also shown to be generalizable across populations and insensitive to imaging apparatus and acquisition pipelines. Frustratingly, much of the research to date has focused on achieving higher accuracy internal classification within datasets rather than considering how these models might generalize to other datasets, and respectively to real world clinical situations. The studies presented in this paper present the results of a large number of experiments testing the generalization ability of deep learning models trained on various publicly available COVID-19 CT image repositories and one private image repository. Although we achieved good to very good results from internal training/testing of these models, external testing displayed more varied results, with our poorest performing models failing to generalize at all, and our best performing models generalizing well.

We were successful in engineering a deep learning model trained on images sourced from Iranian (MID-CT) and Brazilian clinics (SARS-COV-2) and applying these models to predicting COVID-19 lung involvement for an entirely independent dataset sourced from the other side of the world in Moscow with excellent results. Our results demonstrate that a deep learning model trained on a moderately sized, but internally consistent dataset is capable of generalizing to patients from other geographic regions. Our successful models were based on medical image corpuses meeting three simple criteria, being (i) consistent sourcing across image classes, (ii) considered use of histogram equalization/wavelet filtering techniques, and (iii) diversity in patients and CT slice locations. We hope that other researchers can use these principles to guide the development of deep learning models that are generalizable enough to form the basis of useful clinical tools.

Our study does have limitations. We would have preferred to have segmented the lung field from the CT image to improve the signal to noise ratio for the classification experiments. However, our experiments showed that segmentation results using a U-Net (Ronneberger et al., 2015) were highly dependent on the quality of the underlying image, and we were conscious of the possibility that systematically different segmentation artefacts for different image corpuses could potentially impose bias the datasets. To avoid the risk of introducing this bias, we therefore chose to train our model on unsegmented images. In future, having identified that variation in CT slice location appears to promote model generalization, we will procure additional slices for the MID-CT dataset to try to improve upon the results for this dataset.

**Declaration of competing interest**

The authors declare that they have no known competing financial interests or personal relationships that could have appeared to influence the results reported in this paper.